\pdfoutput=1

\documentclass[10pt,twocolumn,letterpaper]{article}

\usepackage{3dv}
\usepackage{epsfig}
\usepackage{times}
\usepackage{graphicx}
\usepackage{amsmath}
\usepackage{amssymb}
\usepackage{comment}
\usepackage{color}
\usepackage{multirow}
\usepackage{textcomp}
\usepackage{tabularx}
\usepackage[font=small]{caption}
\usepackage{subcaption}
\usepackage{makecell}
\usepackage{wrapfig}
\usepackage{breqn}
\usepackage{authblk}

\usepackage{rotating}
\usepackage{cite}

\usepackage[pagebackref=true,breaklinks=true,letterpaper=true,colorlinks,bookmarks=false]{hyperref}

\threedvfinalcopy 

\def\threedvPaperID{151} 

\ifthreedvfinal\fi

\begin{document}

\title{Learning to Guide Local Feature Matches}
\author[1,2]{Fran\c{c}ois Darmon}
\author[2]{Mathieu Aubry}
\author[2]{Pascal Monasse}
\affil[1]{Thales LAS France}
\affil[2]{LIGM (UMR 8049), Ecole des Ponts, Univ. Gustave Eiffel, CNRS, Marne-la-Vall\'ee, France}
\affil[ ]{\tt\small \{francois.darmon, mathieu.aubry, pascal.monasse\}@enpc.fr}

\maketitle

\newcommand{\todo}[1]{\textcolor{red}{TODO: #1}}
\newcommand{\mathieu}[1]{\textcolor{blue}{Mathieu: #1}}

\newcommand{\AB}[0]{ {A \rightarrow B} }
\newcommand{\BA}[0]{ {B \rightarrow A} }
\newcommand{\ii}[0]{ {(i,j)} }
\newcommand{\kk}[0]{ {(k,l)} }
\newcommand{\po}[0]{ \mathcal{P} }
\newcommand{\nega}[0]{ \mathcal{N} }

\maketitle

\begin{abstract}
   We tackle the problem of finding accurate and robust keypoint correspondences between images. We propose a learning-based approach to guide local feature matches via a learned approximate image matching. Our approach can boost the results of SIFT to a level similar to state-of-the-art deep descriptors, such as Superpoint, ContextDesc, or D2-Net and can improve performance for these descriptors. We introduce and study different levels of supervision to learn coarse correspondences. In particular, we show that weak supervision from epipolar geometry leads to performances higher than the stronger but more biased point level supervision and is a clear improvement over weak image level supervision. 
   We demonstrate the benefits of our approach in a variety of conditions by evaluating our guided keypoint correspondences for localization of internet images on the YFCC100M dataset {and indoor images on the SUN3D dataset}, for robust localization on the Aachen day-night benchmark and for 3D reconstruction in challenging conditions using the LTLL historical image data. 
\end{abstract}

\section{Introduction}

\begin{figure}[ht]
    \centering
    \begin{subfigure}[t]{\columnwidth}
        \centering
        \includegraphics[width=0.85 \textwidth, page=1]{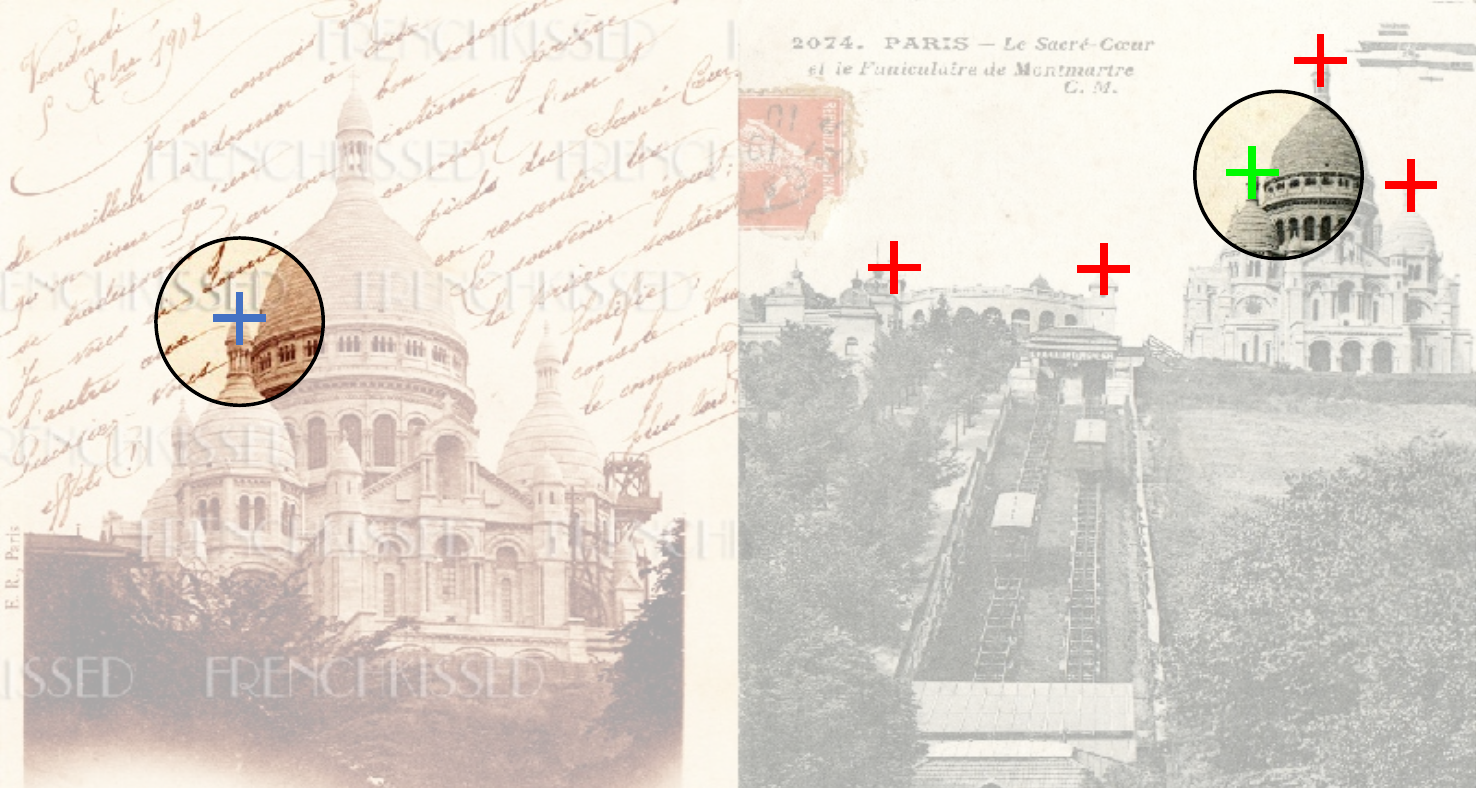}
        \caption{The precise feature match (green) is disambiguated relative to concurrent ones (red) by the coarse matching.}
    \end{subfigure}
    \hfill
    \begin{subfigure}[t]{\columnwidth}
        \centering
        \includegraphics[width=0.85 \textwidth, page=2]{figures/teaser.pdf}
        \caption{SIFT features matched with our method.}

    \end{subfigure}
    \caption{In challenging conditions, local information might not be enough to disambiguate local feature matches. We thus propose to guide the matches using coarse image-level deep correspondences. }
    \label{fig:teaser}
\end{figure}

Image matching is a fundamental task in computer vision and in particular a crucial step of Structure from Motion algorithms. 
Local feature detectors and descriptors are an essential tool for this task, providing both accuracy and high robustness. However, relying exclusively on local information to match images can be misleading 
in particular in the case of repeated or nearly repeated structures. 
We thus propose to complement and guide local keypoint matching using learned image-level coarse correspondences. 

This idea represents an important shift compared to the dominant paradigm where global information and geometric constraints are usually introduced after keypoints have been matched, typically by performing RANSAC~\cite{fischler1981random} to filter matches that are geometrically consistent. Indeed, recent work on applying deep learning to local feature correspondences has mostly focused on improving keypoint detection and description~\cite{dusmanu2019d2, luo2019contextdesc,detone2018superpoint} or improving outlier rejection~\cite{moo2018learning, zhang2019learning}. 
To the best of our knowledge, we are the first to propose a combination of learned coarse correspondences and local keypoint matching, combining the benefits of both approaches.

As illustrated in Figure \ref{fig:teaser} our approach is especially beneficial in challenging conditions and in typical failure cases of classical features. First, when there are repeated structures in the image, they are likely to be disambiguated by the coarse matching and can thus be identified reliably by our guided keypoint matching. Second, when large appearance variations make descriptor matching less reliable, for example in the case of historical images, the number of candidate keypoint matches is reduced strongly by our method and false matches are less likely to appear.

We demonstrate that our approach boosts the results obtained with the standard SIFT descriptor to a level similar to the most advanced state-of-the-art deep descriptors. Our method can also be used with more advanced detectors and descriptors and we demonstrate it also boosts their performance, though by a smaller margin. This is a hint that a large part of the improvement brought by modern deep keypoint descriptors comes actually from their ability to consider global image information instead of exclusively at the local level. Note that this is explicitly targeted in some approaches and deep architectures such as ContextDesc~\cite{luo2019contextdesc}.




The main challenge for guided matching is to predict coarse image correspondences. We build on an architecture computing correlation between base deep features and filtering them using a 4D convolutional network~\cite{rocco2018neighbourhood}. This approach has the advantage to be able to handle any displacement and to leverage geometric consistency via the 4D convolutions. It is possible to train it with only weak image supervision, providing the network with matching and non-matching image pairs. We introduce and study two other levels of supervision: weak epipolar supervision and point supervision. Indeed 
we can exploit large scale databases of 3D models reconstructed via Structure from Motion~\cite{heinly2015reconstructing, li2018megadepth} that provide camera calibration, from which we can infer epipolar constraints for all points as well as a sparse set of reconstructed points that can be used as ground truth matches. This data is, of course noisy and biased since points could only be reconstructed when traditional approaches succeeded, but we demonstrate it can still be used to boost performances.
Contrary to~\cite{rocco2018neighbourhood}, both the weak epipolar supervision and point supervision improved results by fine-tuning the base features.
\\
Our three main contributions are the following:
\begin{enumerate}
    \item We propose the first learned guided correspondence approach for local keypoint matching. 
    
    \item We study different possible levels of supervision to learn coarse image matching, in particular weak supervision from epipolar geometry.
    
    \item We demonstrate our method benefits all the studied keypoint descriptors. In some cases, it boosts the traditional SIFT descriptor to the performance of the latest learned descriptors, hinting it is mainly due to their discriminating power by considering global image characteristics. 
\end{enumerate}

\section{Related Work}
\noindent \textbf{Local features.}
There exist many local feature detectors and descriptors~\cite{mikolajczyk2005comparison,mikolajczyk2005performance}, SIFT~\cite{lowe2004distinctive} being likely the most known and used. Recently, deep learning based methods have gained popularity. Geodesc~\cite{luo2018geodesc}, ContextDesc~\cite{luo2019contextdesc}, HardNet++~\cite{mishchuk2017working} and HesAffNet~\cite{mishkin2018repeatability} describe pre-extracted patches using a neural network with different training procedures. LogPolarDesc~\cite{ebel2019beyond} introduces a new patch  representation more adapted to neural networks. In LIFT~\cite{yi2016lift}, LF-net~\cite{ono2018lf}, SuperPoint~\cite{detone2018superpoint}, D2-net~\cite{dusmanu2019d2} and R2D2~\cite{revaud2019r2d2} both extraction and description are learned. Our approach can be used to match any of these local features. High level semantic information can be learned by some descriptors~\cite{detone2018superpoint, luo2019contextdesc, dusmanu2019d2, revaud2019r2d2}; our experiments indicate that even with these descriptors our coarse image correspondence guidance can lead to better performance.

\noindent \textbf{Spatial verification.}
Classical image matching pipelines perform keypoint matching then correspondence pruning using a ratio test~\cite{lowe2004distinctive} or bidirectionnal check, allowing to remove the ambiguous matches. More elaborate techniques like CODE~\cite{lin2017code}, GMS~\cite{bian2017gms} and LPM~\cite{ma2019locality} further remove false matches with the observation that keypoint matches should be consistent with their close neighbors. Then a robust estimator is used for geometry estimation, the most widely used being RANSAC~\cite{fischler1981random}. Recent approaches~\cite{moo2018learning, ranftl2018deep, zhang2019learning, brachmann2019neural} learn outlier filtering by neural networks. They typically consider the matches as a 4D point cloud. These point cloud networks can be supervised with epipolar geometry~\cite{hartley2003multiple}: if the fundamental matrix between two images is available, each match can be assigned a label as inlier or outlier depending on its epipolar distance. However, all these correspondence pruning techniques and robust estimators cannot correct, but only discard, wrong matches. On the contrary, our approach leverages spatial information before the matching step and can help to identify correct matches. 

\noindent \textbf{Guided matching.}
Several works \cite{hartley2003multiple, feng2011feature, shah2015geometry, maier2016guided} introduced the idea of using an existing geometric model to guide keypoint matches. \cite{hartley2003multiple} proposes to use a homography model, \cite{shah2015geometry} a fundamental matrix model, \cite{feng2011feature} a combination of both and \cite{maier2016guided} a specifically designed keypoint-based statistical optical flow. However, all these methods require an accurate initial keypoint based estimation of the geometric model in order to get more keypoint matches. For challenging scenarios such as day-night matching this is not realistic and adding guided matches from an incorrect geometry would add even more false matches. Other approaches \cite{widya2018structure, taira2019inloc} match features of a pre-trained CNN in a hierarchical manner by first matching coarse deep features then higher resolutions features inside the receptive field of the matched features. Although very intuitive, it also requires good initial matches and we show that using pre-trained CNN features does not lead to good matches. 

\noindent \textbf{Learned matching.}
Independently from keypoints, Deep Learning can be applied to image matching. It was first applied on optical flow \cite{dosovitskiy2015flownet, ilg2017flownet, ranjan2017optical, sun2018pwc} and homography estimation \cite{detone2016deep}. However, those methods are not able to handle large geometry variation. Rocco et al.~\cite{rocco2017convolutional, rocco2018end} deal with this issue by using global image transformation models such as affine transformation or thin-plate spline but such models are often not relevant for 3D scenes. Both \cite{sun2018pwc,melekhov2019dgc} try to remedy this problem by using a coarse to fine or iterative approach. Neighborhood Consensus Network~\cite{rocco2018neighbourhood} proposes to use a 4D convolution network without any prior image transformation model. Recently, SuperGlue \cite{sarlin2020superglue} introduced a graph neural network that learns to match local features. 

\section{Guided Feature Matching}
\begin{figure}
    \centering
    \includegraphics[width=\columnwidth]{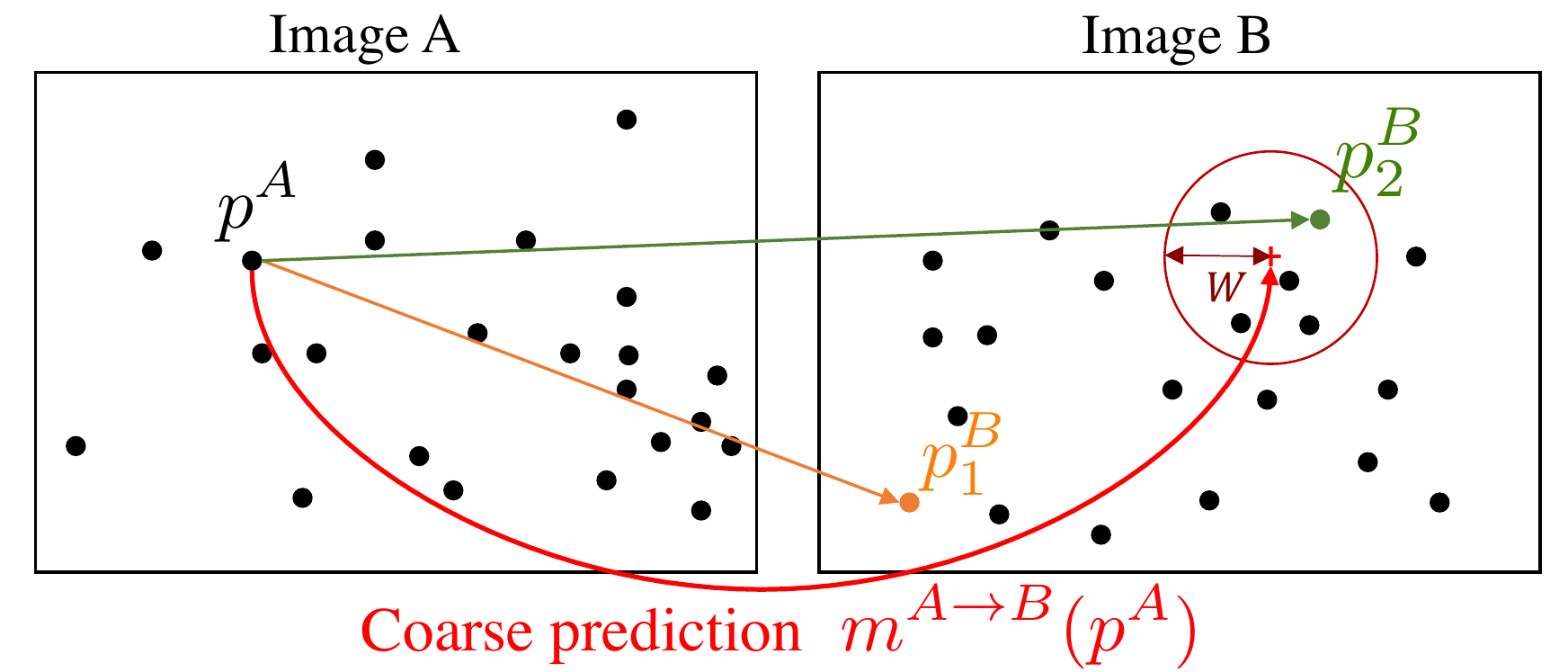}
    \caption{Guiding keypoint matching with a coarse match. The orange match $p^B_1$ is the closest in term of descriptor distance, but is not consistent with the coarse prediction. The correctly selected match, $p^B_2$ (in green), is the closest in descriptor space being consistent with the coarse prediction.}
    \label{fig:guided_matching}
\end{figure}

Local keypoints have clear advantages for robust image matching. Indeed, they are naturally robust to occlusion of part of the image, localized changes, and clutter. Keypoint detectors are also designed to localize points with sub-pixel accuracy and to be robust to changes of viewpoint. However, local image regions are insufficient to reliably match keypoints in the presence of repetitive structures, which only large scale image information can help disambiguate. More generally, matches have to be identified among all the keypoints in the target image, and thus good matches have to be distinguished from a large number of false correspondences.
We propose to make keypoint matching easier by first using a neural network to predict coarse correspondences at image level, and using them to guide keypoint matching, considering candidate matches only in a small image region.

This idea is illustrated in Figure~\ref{fig:guided_matching}. 
Let us assume we have access to an approximate match $m^\AB$ between images~$A$ and~$B$. We want to match a keypoint at position $p^A$ in image~$A$, described by a feature $f^A$ to the keypoints detected in image $B$ at positions $p^B_i$, described by features $f^B_i$, for $i = 1 \dots N$. We will leverage  $m^\AB$ by comparing $f^A$ only to features of keypoints close to its approximate match $m^\AB(p^A)$. The index~$j$ of the optimal match is given by:



\begin{equation}
    j = \underset{i : \|m^\AB(p^A) - p^B_i\| <W}{\arg\min}\|f^A - f^B_i\|,
\end{equation}
 where $W>0$ is a parameter of our method. Note that using $W=\infty$ leads back to the standard matching.
Similarly, the matching can be performed from image B to image A and the mutual matching test can remove outliers. 

\section{Learning coarse correspondences}

\begin{figure*}[!t]
    \centering
    \includegraphics[width=\textwidth]{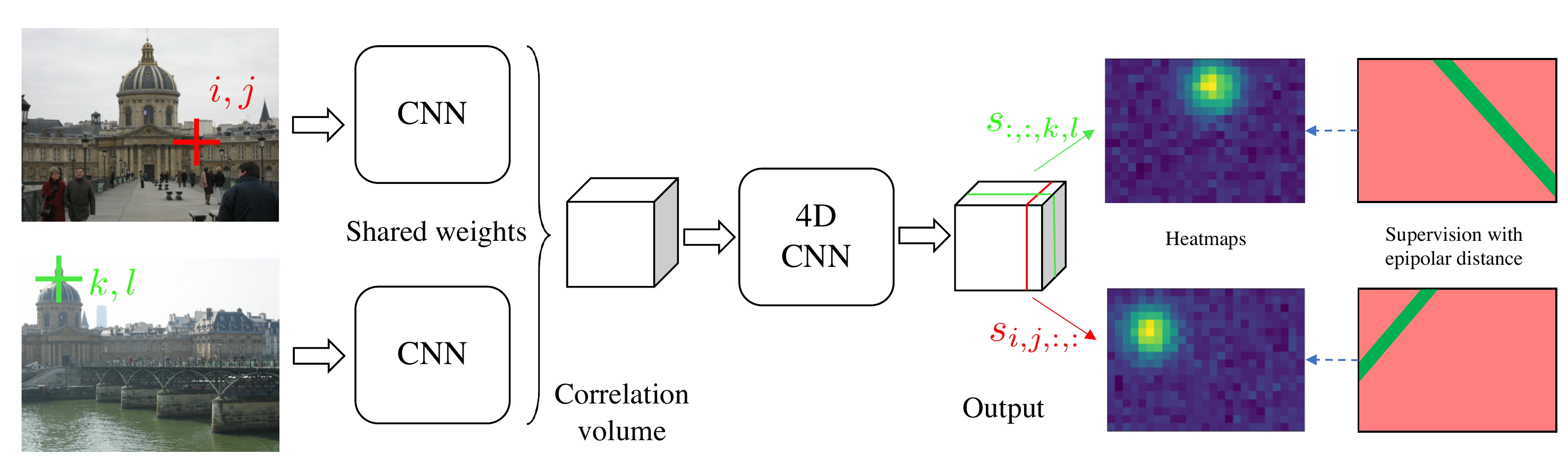}
    \caption{Overview of our coarse matching network architecture and training: a shared CNN extracts coarse features from the two images. These features are then combined via the dot product into a 4D correlation volume. This volume is finally filtered with a CNN based on 4D convolutions, which can be trained with weak epipolar supervision.}
    \label{fig:architecture}
\end{figure*}

In this section, we present our deep learning approach to predict approximate correspondences between images. The key elements of our approach are visualized in Figure~\ref{fig:architecture}. 
In the following, we first discuss our architecture, then present losses corresponding to three levels of supervision, and finally provide details of our implementation and training. 

\subsection{Architecture}

We build on the NCNet~\cite{rocco2018neighbourhood} architecture. We first compute feature maps $f^A$ and $f^B$ for both input images and 
aggregate them in a 4D correlation volume $c_{ijkl} = \langle f^A_{ij} | f^B_{kl} \rangle$ that contains the correlation between every feature in image~$A$ and every feature in image~$B$. We use a 4D convolutional neural network to filter the correlation volume into a new volume~$s$, trained to have high values only in positions corresponding to valid correspondences.  \cite{rocco2018neighbourhood} motivates this architecture and the use of 4D convolutions by the idea of neighborhood consensus: the quality of a match between feature $\ii$ in image~$A$ and $\kk$ in image~$B$ should be decided not only based on the correlation $c_{ijkl}$ but also on the correlation of the neighbor features. 

The coarse matches between features in each direction are extracted from $s$ using an argmax over the target image's dimensions. Such matches can be interpolated at pixel level: given a point in pixel coordinate $(x,y)$ in image~$A$, its coarse match  $m^\AB(x,y)$ in image~$B$ is computed using  bilinear interpolation of the feature matches of its four nearest features.  Inversely, $m^\BA(x,y)$ denotes the coarse match in image~$A$ of pixel $(x,y)$ in image~$B$.

\subsection{Supervision}

We now introduce three different levels of supervision corresponding to different information about the ground truth matches and the associated losses. First, we consider an image level supervision, given in the form of pairs of matching and non matching images. Second, we introduce an epipolar supervision, which in addition leverages geometry information to infer a line where positive matches can lie. Third, we discuss a loss for point supervision, which uses ground truth matches between images. 

In the rest of the section, we assume we are given a set $\mathcal{T}$ of training image pairs $(A_m,B_m)_{m=1...M}$ and we minimize the loss:
\begin{equation}
    \mathcal{L}=\sum_{(A,B)\in\mathcal{T}} l^\AB(s^\AB) + l^\BA(s^\BA)
\end{equation}
where  $s^\AB$ (resp. $s^\BA$) is the result of applying a softmax to $s$ in the dimensions corresponding to image $B$ (resp. $A$) and $l^\AB$ and $l^\BA$ are the losses associated to the matches in both directions. For simplicity we only explain $l^\AB(s^\AB)$ in the following subsections.


\subsubsection{Weak image level supervision}

For image level supervision, we use the same loss as~\cite{rocco2018neighbourhood}. For each pair $(A,B)$ of images, we write $y_{AB}=1$ if both images represent the same scene and $y_{AB}=-1$ otherwise. We then define the loss by: 

\begin{equation}
    l^\AB_{image} = -y_{AB}\sum_{\ii}\max_{\kk} s^\AB_{ijkl}.
\end{equation}
This loss encourages the maxima of  $s^\AB$ to be 1 for as many features as possible when the image pair is positive, which amounts to making the maxima in~$s$ sharper, and on the contrary when the pair is negative encourages the maxima of $s^\AB$ to be small, which amounts to having almost constant values in~$s$. In order to balance the influence of negative and positive examples, the training batch consists of one half positive and one half negative image pairs. This supervision has been shown to achieve good results for semantic matching when image pair label is typically the only supervision available. However, we argue that additional information provided by 3D reconstruction datasets improve the matches. 

\subsubsection{Weak epipolar supervision}

We propose to leverage epipolar geometry~\cite{hartley2003multiple} to better supervise the matches. Given a position $\ii$ in image~$A$, it is possible to use the camera calibrations (internal parameters and 6D pose) to predict the epipolar line on which the corresponding point in image~$B$ will lie. The distance between a position $\kk$ in image~$B$ and this line is called the epipolar distance
\begin{equation}
    d^F(\ii, \kk) = \frac{|(k,l,1) F (i,j,1)^\top|}{\sqrt{(F(i,j,1)^\top)_{[1]}^2 + (F(i,j,1)^\top)_{[2]}^2}},
\end{equation}
\noindent where $\mathbf{t}_{[i]}$ denotes the $i$th coordinate of vector $\mathbf{t}$ and $F$~is the fundamental matrix associated to the image pair, computed from the full calibration.
We design a loss to leverage this information. 
Instead of trying to increase all maxima in positive image pairs, we try to increase only the ones consistent with epipolar geometry.  
Let $\mathcal{P}^\AB$ be the subset of features in image~$A$ whose matches are consistent with epipolar geometry,
\begin{equation}
    \mathcal{P}^{\AB} = \left\{(i,j) |  d^F\left((i,j), \underset{(k,l)}{\text{argmax }} s^\AB_{ijkl}\right) < \lambda \right\},
\end{equation}
\noindent where $\lambda$ is a threshold on the epipolar distance, and $\mathcal{N}^\AB$ the complementary set of $\mathcal{P}^\AB$, which correspond to matches that are not consistent with epipolar geometry. We propose to use as loss:
\begin{dmath}
    l^\AB_{epipolar} = \dfrac{1}{2 |\mathcal{N}^\AB|}\sum_{(i,j) \in \mathcal{N}^\AB}\max_{kl} s^\AB_{ijkl}
    - \\ \dfrac{1}{|\mathcal{P}^\AB|}\sum_{(i,j) \in \mathcal{P}^\AB}\max_{kl} s^\AB_{ijkl}
\end{dmath}
As in the previous section, we use images from different scenes for half the batch. We consider that all the points for such image pairs are in $\mathcal{N}^\AB$ and that the second term is zero. The division by 2 of the negative part of the loss then balances the positive and negative parts. 


%
%

\subsubsection{Point supervision}

Point supervision is the strongest form of supervision we consider. It relies on sparse ground truth match labels. Let us assume that we are given a set of~$N$ ground truth correspondences between images $(p_1^A, p_1^B) \dots (p_N^A, p_N^B)$. Let $\mathcal{M}^\AB(i,j)$ be the set of features in image~$B$ that have a ground truth match with feature at $(i,j)$ in image~$A$. The loss we use for point supervision is

\begin{equation}
    l^\AB_{points} = - \sum_{ij} \max_{(k,l) \in \mathcal{M}^\AB(i,j)} s^\AB_{ijkl}.
\end{equation}
This loss simply encourages $s^\AB$ to be as close to~1 as possible for the best corresponding feature.  
Note that we could also use negative contributions as for the image level and epipolar supervision, or inversely consider only positive contributions for the epipolar supervision. 
We experimented with these variations and found that they lead to results worse than the losses we have discussed.

\subsection{Implementation and training details}

Similar to D2-Net\cite{dusmanu2019d2} we train the coarse matching network on MegaDepth dataset~\cite{li2018megadepth}. This dataset consists of 196 sets of images collected from the same physical scene. COLMAP \cite{schonberger2016structure} was run on these scenes to obtain a sparse 3D reconstruction. 
We removed from the training set all the scenes that are used in the evaluation: 
the Tanks and Temples scenes from FM benchmark~\cite{bian2019bench}, the 4 YFCC scenes \cite{thomee2016yfcc100m, heinly2015reconstructing} evaluated in OANet~\cite{zhang2019learning}, the 6 YFCC scenes of Image Matching Workshop \cite{imw} and the buildings from LTLL~\cite{fernando2015location}.
This reduces the training set to 175 scenes. We use the provided calibration for our weak supervision and choose as positive image pairs the ones that see at least $30$ common 3D points in the reconstruction.

We use Resnet101 \cite{he2016deep} Conv4 features pretrained on ImageNet to extract feature maps from the input images. The 4D CNN is composed of three successive 4D convolutions layers with 16 channels and kernel of size 3. Similar to NCNet we ensure the output volume is independent of the image order by feeding the images in both orders successively and by taking the average of the outputs. The networks are trained with the Adam optimizer, an initial learning rate of $10^{-3}$, and a batch size of $8$ for $25000$ iterations. For the epipolar supervision, $\lambda$ is set to the distance between two consecutive features. The networks are initially trained with frozen feature extractors. Then after convergence, the feature extractors can be fine-tuned with a smaller learning rate. We limit the image resolution at $401$ pixels at training time and keep the original aspect ratio with zero padding. At test time we limit the resolution to $497$, which gives a feature resolution of at most $32 \times 32$. For a typical $1600\times1600$ image, each feature will correspond approximately to a $50$ pixels square.

\section{Experiments}

In this section, we compare our approach with other guided matching methods, correspondence filtering techniques and state of the art features. First we validate and analyse the performance of our coarse matching network. Second, we compare our approach to other guided matching and correspondence pruning techniques.
Third, we use our guided matching with different keypoint detectors and descriptors and show that our method consistently improves their results.  Finally, we show that our method can help 3D reconstruction on challenging scenes.

\subsection{Coarse matching}

\begin{table}[t]
    \centering
    \begin{tabular}{c|c c c|c c c}
         & \multicolumn{3}{c|}{Frozen features} & \multicolumn{3}{c}{Finetuned features} \\
        Threshold &  8 & 16 & 32 & 8 & 16 & 32 \\
        \hline
        Image \cite{rocco2018neighbourhood} & 34.5 & 55.0 & 65.36 & 36.3 & 57.8 & 68.7\\
        Epipolar &  43.1 & 62.4 & 70.7 & \bf 47.7 & \bf 67.6 & \bf 75.8\\
        Point & 40.3 & 58.5 & 67.8 & 45.0 & 63.5 & 72.5 \\
    \end{tabular}
    \caption{Proportion of ground truth SfM points from MegaDepth correctly predicted by the coarse matcher. The threshold is in pixel units in the resized image coordinates. Here, 16 pixels is the distance between two consecutive features. Guiding with epipolar supervision leads to the highest proportion of matches in the guidance.}
    \label{tab:coarse_matching_pck}
\end{table}

\begin{table*}[ht]
    \centering
    \begin{tabular}{c|c|c c c| c c c}
        \multirow{2}{*}{Matches} & \multirow{2}{*}{Pre-filtering}& \multicolumn{3}{c|}{YFCC (internet)} & \multicolumn{3}{c}{Sun3D (indoor)} \\
        &&  5\textdegree & 10\textdegree & 20\textdegree & 5\textdegree & 10\textdegree & 20\textdegree \\
        \hline
        \multirow{5}{*}{Raw} & None & 8.45 & 13.80 & 22.4 & 2.34 & 4.70 & 9.61 \\
         & Bidirectional check & 27.70 & 36.43 & 47.73 & 6.96 & 11.72 & 19.89 \\
         & Ratio test & 41.75 & 51.63 & 62.23 & 13.48 & 20.93 & 31.48 \\
         & Ratio test + bid. check & 46.80 & 57.41 & 67.80 & 14.52 & 22.74 & 34.22 \\

         & Ratio test + GMS \cite{bian2017gms} & 30.43 & 38.30 & 48.16 & 11.49 & 17.89 & 27.46 \\
        \hline
        \multirow{4}{*}{Raw} & CNNet \cite{moo2018learning,zhang2019learning} & 47.98 & 58.13 & 68.67 & 15.98 & - & - \\
        & N$^3$Net \cite{plotz2018neural, zhang2019learning} & 49.13 & - & - & 15.38 & - & -\\
        & DFE \cite{ranftl2018deep, zhang2019learning} & 49.45 & - & - & 16.45 & - & -\\
        & OANet \cite{zhang2019learning} & \bf 52.08  & \bf 62.38 & \bf 72.66 & \bf 17.25 & \bf 26.60 & \bf 39.50 \\
        \hline
        Guided epipolar~\cite{shah2015geometry} & Ratio test + bid. check & 45.88 & 55.59 & 65.20 & 15.86 & 24.52 & 36.31 \\
        Guided homography & Ratio test + bid. check & 46.00 & 55.65 & 65.46 & 15.15 & 23.55 & 35.36\\
        Guided VGG4 \cite{widya2018structure, taira2019inloc} & Ratio test + bid. check & 31.23 & 40.49 & 51.51 & 3.97 & 7.23 & 13.16 \\
        \hline
        Ours image guided & Ratio test + bid. check & 43.50 & 52.99 & 63.24 & 15.45 & 23.84 & 35.81 \\
        Ours point guided & Ratio test + bid. check & 47.43 & 57.71 & 68.59 & 15.61 & 24.24 & 36.37 \\
        Ours epipolar guided & Ratio test + bid. check & 49.60 & 60.36 & 71.37 & 15.72 & 24.35 & 36.40 \\

    \end{tabular}
    \caption{Comparison with various correspondence filtering and guided matching  methods on 2-view geometry estimation. We report the AUC for a given tolerance for rotation and translation direction. The matches are computed from 2000 SIFT keypoints. ``Ours Image/Point/Epipolar guided'' is our guided matching with the different supervisions. A final RANSAC filtering follows any used pre-filtering.}
    \label{tab:filtering_comparison}
\end{table*}

We first evaluate our coarse matching using the 3D points provided by MegaDepth as ground truth matches for a set of $1600$ test image pairs. 
For each ground truth match $(p^A, p^B)$, we compute the distance $\| m^\AB (p^A)-p^B\|$. The proportion of distances below a threshold is used for evaluation. We use as threshold $8$, $16$ and $32$ pixels since the distance between two nearby coarse matches is $16$ pixels. 

We report in Table~\ref{tab:coarse_matching_pck} the results obtained with our different supervisions as well as fine-tuning or not the ResNet-101 feature extractor, which was reported to degrade performances in the test database of~\cite{rocco2018neighbourhood}. However, in our experiments finetuning the feature extractor leads to better matching, its effect being stronger with the epipolar and point supervisions. As can be expected, image supervision leads to the worst results. Although it is trained with a stronger supervision, point supervision has worse performances than epipolar supervision. This may be because point supervision is sparse and biased, providing information on specific areas of the image only.  
With a window size $W = 16$, the performance of epipolar supervision is close to $70\%$, which seems acceptable for guiding keypoint matching; we use this threshold to filter our matches in the rest of the experiments. 

\begin{figure*}[!ht]
    \centering
    \includegraphics[width=\linewidth]{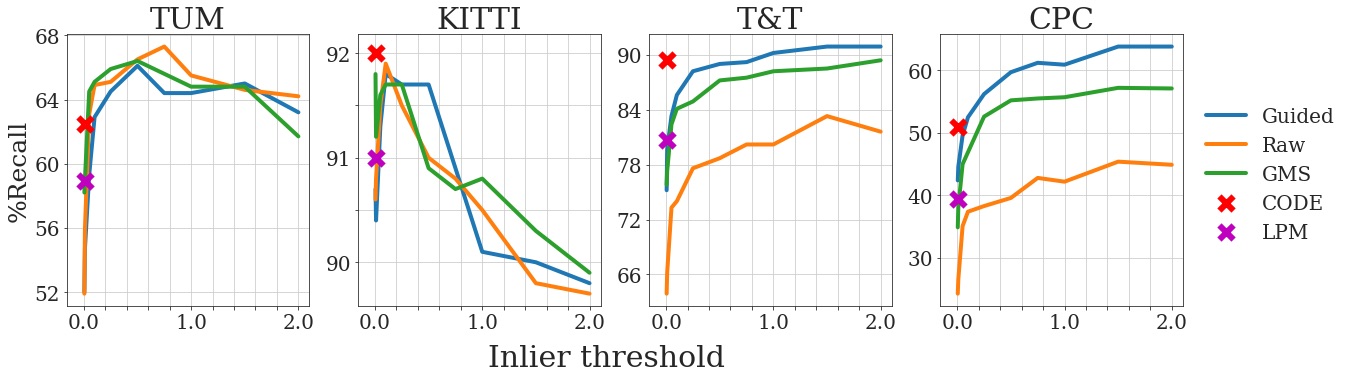}
    \caption{FM benchmark results for a varying RANSAC threshold. A different RANSAC threshold must be carefully chosen for every dataset and every method for fair comparison. The compared methods perform similarly on KITTI and TUM but our guided matching performs the best on the wide baseline datasets. }
    \label{fig:fm_benchmark}
\end{figure*}

\begin{table*}[ht]
    \centering
    \begin{tabularx}{\linewidth}{c c|X X X|X X X|X X X}
        \multirow{2}{*}{Features} & \multirow{2}{*}{Matching} & \multicolumn{3}{c|}{\thead{YFCC two-view \\geometry estimation}}& \multicolumn{3}{c|}{\thead{Sun3D two-view \\geometry estimation}} & \multicolumn{3}{c}{\thead{Aachen day/night\\ visual localization}} \\
        & & 5\textdegree & 10\textdegree & 20\textdegree & 5\textdegree & 10\textdegree & 20\textdegree & (0.25m, 2\textdegree) & (0.5m, 5\textdegree) & (5m, 10\textdegree)\\
        \hline
        \multirow{2}{*}{\thead{SIFT}} 
        & Raw &  46.80 & 57.41 & 67.80 & 14.52 & 22.74 & 34.22 & 38.8 & 51.0 & 58.2 \\
        & Ours Epip. & \bf 49.60 & \bf 60.36 & \bf 71.37 & \bf 15.72 & \bf 24.35 & \bf 36.40 & \bf 66.3 & \textbf{84.7} & \bf 96.9\\ \hline
        \multirow{2}{*}{\thead{Contex- \\xtDesc}} 
        & Raw & \bf 55.40 & \bf 66.58 & \bf 77.38 & \bf 16.83 & \bf 25.77 & \bf 37.99 & 60.2 & 74.5 & 87.8 \\
        & Ours Epip. & 51.95 & 62.60 & 73.33 & 16.50 & 25.43 & 37.56 & \textbf{75.5} & \bf 85.7 & \bf98.0\\
        \hline
        \multirow{2}{*}{\thead{Super- \\point}} 
        & Raw & 32.48 & 42.84 & 54.25 & 15.39 & \bf 24.27 & \bf 36.37 & 70.4 & 77.6 & 85.7 \\
        & Ours Epip. & \textbf{38.10} & \textbf{49.06} & \textbf{61.48} & \textbf{15.60} & 24.23 & 36.33 & \bf 75.5 & \textbf{89.8} & \bf 99.0\\ 

        \hline
        \multirow{2}{*}{D2-Net}
        & Raw & \bf 25.20 & \bf 35.63 & 49.43 & 13.52 & 22.67 & 35.61 & \bf 78.6 & 85.7 & \textbf{100} \\
        & Ours Epip. & 24.68 & 35.30 & \bf 49.55 & \bf 14.10 & \bf 22.87 & \bf 35.63 & 76.5 & \bf 87.8 & 99.0\\ 
        
    \end{tabularx}
    \caption{Comparison with state of the art keypoint detectors and descriptors. We report AUC on several localization thresholds for YFCC and Sun3D, and the proportion of image sucessfully localized for Aachen benchmark. Raw descriptor denotes classical matching with mutual test and RANSAC. Ours Epip. is our guided matching with epipolar supervisions, mutual test, and RANSAC. We only show the best results for several ratio test thresholds (including no ratio test at all) before the other outlier filtering steps. Note that D2-Net's training set intersects YFCC100M test set.}
    \label{tab:pose_estimation}
\end{table*}

\subsection{Comparison with guided matching and correspondence pruning}

There is no direct benchmark for sparse matching. However, as mentioned earlier, sparse matching is the backbone of many 3D related tasks for which datasets exist and allow to indirectly evaluate the quality of matches. We compare our method for matching SIFT features with a posteriori filtering techniques and traditional guided matching on 2-view geometry estimation, both outdoor and indoor. 

First, we use the setup of~\cite{zhang2019learning} to evaluate 2-view geometry accuracy on pairs of images from the YFCC100M and Sun3D datasets. The YFCC100M dataset~\cite{thomee2016yfcc100m} is a very large collection of internet images that was used for Structure from Motion in~\cite{heinly2015reconstructing}. Four scenes and 1000 image pairs per scene are used for evaluation.  Sun3D~\cite{xiao2013sun3d} data come from RGBD indoor videos. $15$ indoor scenes and 1000 image pairs per scene are used for the evaluation. 
On both datasets, for each image pair, the matches provided by different approaches are used to estimate the essential matrix with RANSAC, which in turn is used to compute the relative pose (rotation and translation)~\cite{hartley2003multiple}.

Our results are reported in Table~\ref{tab:filtering_comparison}. We compare our method for matching 2000 SIFT features with several correspondence pruning methods after classical nearest neighbor matching (raw matching). We also report the results for traditional guided matching baselines. Following \cite{shah2015geometry} the top 20\% features in term of scale are first matched in order to estimate a geometric model. The model is then used to guide feature matching. We evaluate two geometric models: homography and fundamental matrix~\cite{shah2015geometry}. 
We also compare with the pretrained VGG4 guided matching of~\cite{widya2018structure, taira2019inloc}: for each mutual match between VGG4 features, we match the SIFT features located inside the receptive field of the corresponding VGG features. For clarity purpose, we only report for the ratio test experiments the results with the ratio that performed the best among $0.8$, $0.9$ and $0.95$. More details can be found in supplementary material. 
Our method ranks second for two view geometry estimation after OANet. Interestingly, as noted in~\cite{sun2019attentive}, the ratio test is very important for SIFT matching; combined with bidirectional check, it is a very strong baseline. We note again that the epipolar supervision performs clearly better than the point supervision. 

Second, we evaluate on the FM Benchmark~\cite{bian2019bench}, a combination of scenes of Tanks and Temple (T\&T)~\cite{knapitsch2017tanks}, TUM~\cite{sturm2012benchmark}, KITTI~\cite{geiger2012we} and Community Photo Collection (CPC)~\cite{wilson2014robust} datasets. Similar to the previous setup, the sparse matches are used to estimate the fundamental matrix that is compared with the ground truth. Each method is compared using the recall: the proportion of fundamental matrices correctly estimated. This metric is very sensitive to the inlier threshold chosen for RANSAC so we show in Figure~\ref{fig:fm_benchmark} the recall of raw matches, our method and GMS~\cite{bian2017gms} for various inlier thresholds. We also show results of the benchmark at the default threshold of $0.01$ for CODE~\cite{lin2017code} and LPM~\cite{ma2019locality}. Since TUM dataset is an indoor dataset with short baseline, the difficulty lies more in the keypoint detection than on the matching and it is not surprising that all methods provide similar results. For KITTI, the results seem saturated and every method also performs similarly. On the two wide baseline scenes, our method shows a large improvement on raw SIFT matching and outperforms GMS by a significant margin. 

\subsection{Validation on learned keypoint detectors and descriptors}

\begin{figure*}[!ht]
    \begin{subfigure}[b]{0.37 \linewidth}
        \includegraphics[width=\textwidth]{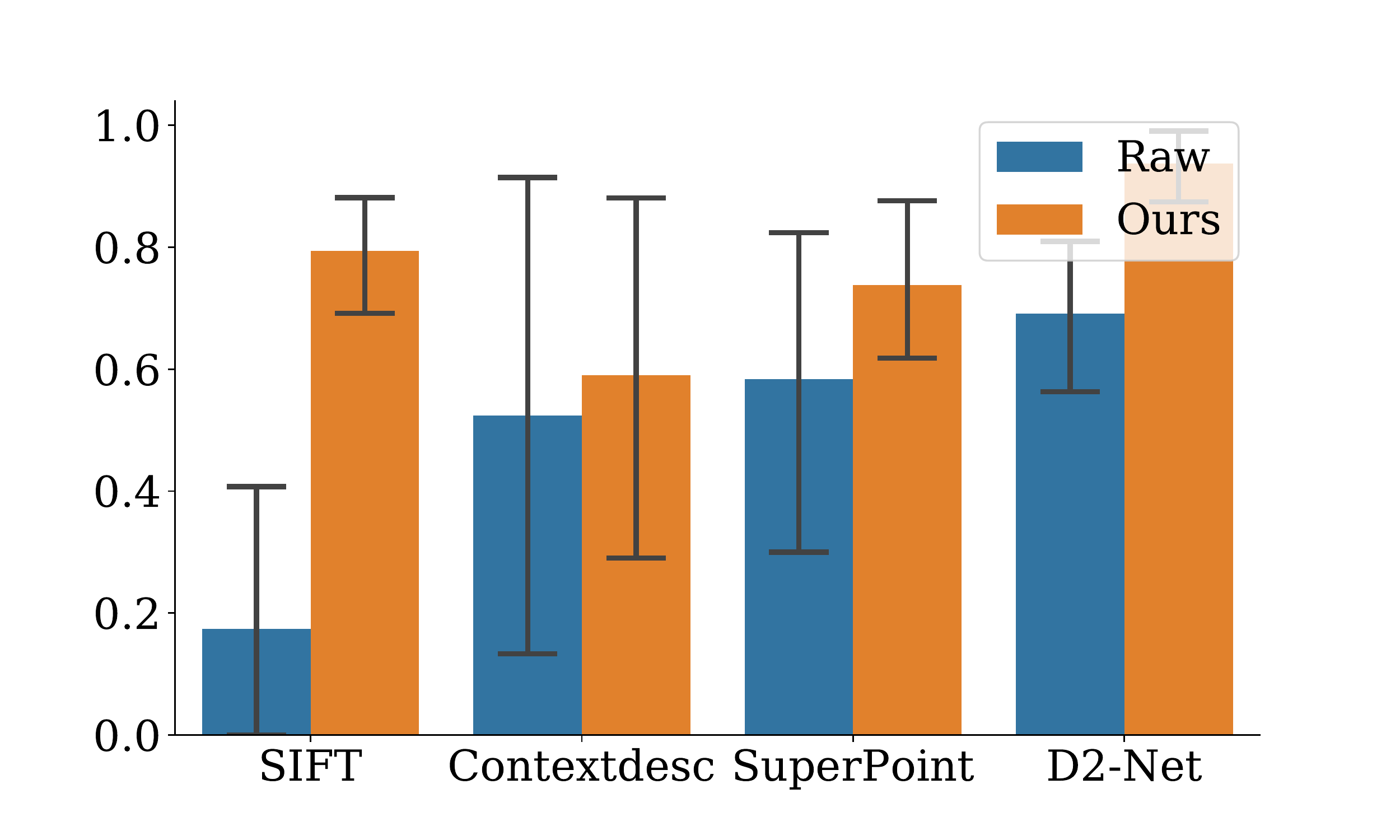}
        \caption{Proportion of image registered}
        \label{fig:prop_registered}
    \end{subfigure}
    \begin{subfigure}[b]{0.31\linewidth}
    \centering
        \includegraphics[height=48pt]{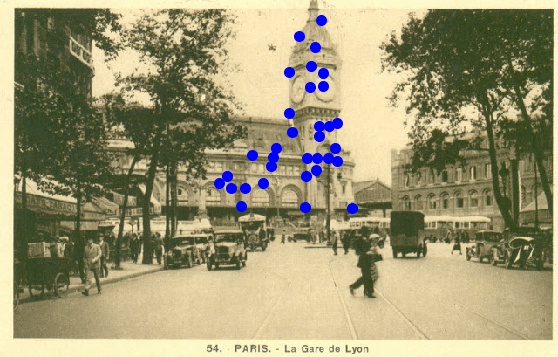}
        \includegraphics[height=48pt]{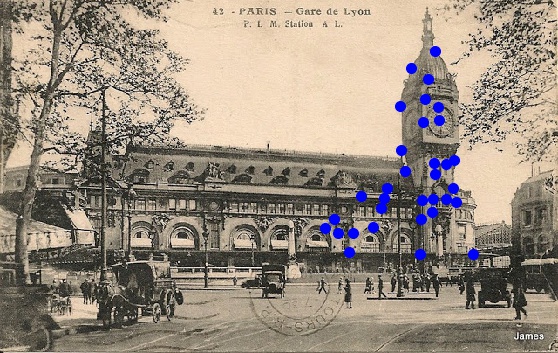}
        \includegraphics[height=48pt]{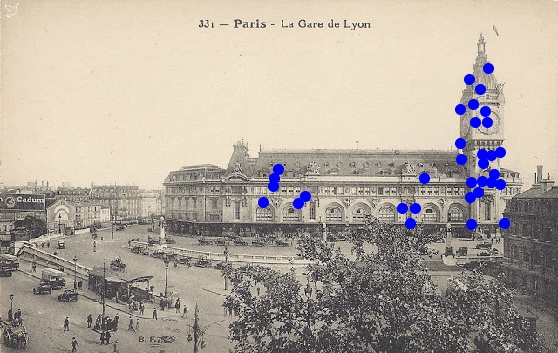}
    
        \caption{SuperPoint raw descriptor}
    \end{subfigure}
    \begin{subfigure}[b]{0.31 \linewidth}
        \centering
        \includegraphics[height=48pt]{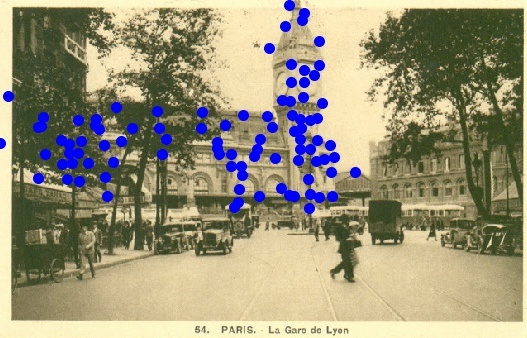}
        \includegraphics[height=48pt]{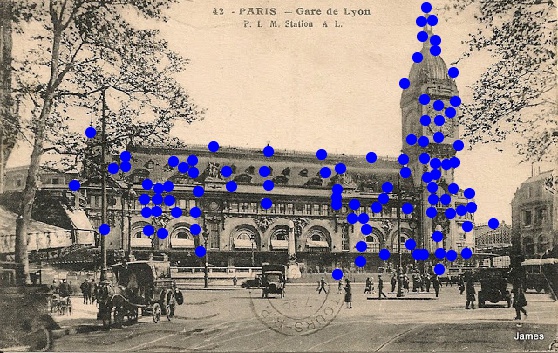}
        \includegraphics[height=48pt]{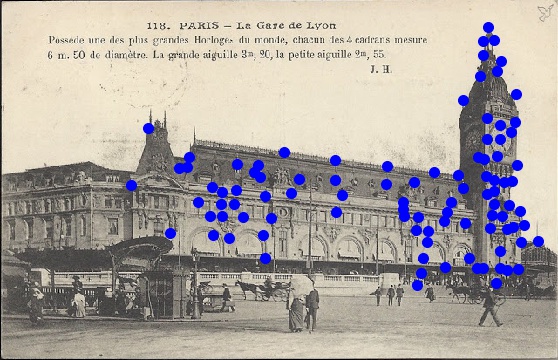}
        \includegraphics[height=48pt]{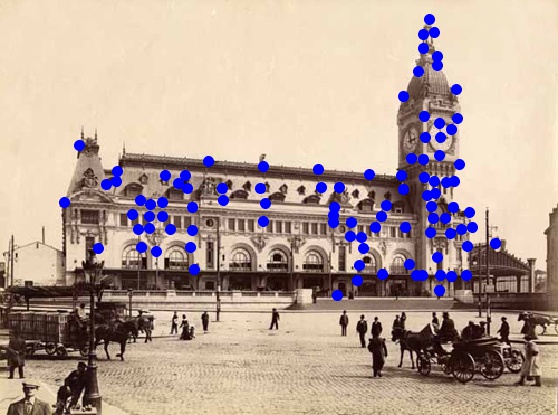}
        \caption{SuperPoint with our guided matching }
    \end{subfigure}
    
    \caption{3D reconstruction results on challenging image sets~\cite{fernando2015location}. We show the average proportion of image registered in each scene of the historic sets in (a) and  a comparison of the 3D reconstructions of scene ``Sacre Coeur'' for Superpoint raw descriptor (b) and our method (c). We show the projection of the reconstructed point cloud on every image added into the model. Our matching not only helps incorporate more images in the reconstruction but it also make it more dense.}
    \label{fig:rec3D}
\end{figure*}

In this section, we show that our guided matching benefits to many features, including the most recent learned deep features, by improving their results for 2-view geometry estimation and visual localization. In addition to the two-view geometry estimation results on YFCC100M and SUN3D, we report visual localization results on the local feature challenge from the Aachen day/night benchmark~\cite{sattler2018benchmarking}. 
The challenge provides a list of image pairs to match, from daytime to daytime, and from nighttime to daytime. The daytime to daytime matches are used to build a 3D point cloud. Then the nighttime to daytime matches are used to register the nighttime images to this model. The evaluation measure is the mean average precision (mAP) of the localization of all the query nighttime images. Note that since the evaluation is performed on 98 images only, small differences in performance should not be over-interpreted.

Our results on Aachen daynight as well as YFCC100M and SUN3D are reported in Table~\ref{tab:pose_estimation}. Most features are improved by our method. Our guidance does not benefit ContextDesc on YFCC and Sun3D, hinting that this method is effective in adding global context. However, its performance is still improved on the harder Aachen day/night benchmark. The results for D2-net are inconclusive on the Aachen day/night benchmark but improvements are visible on SUN3D. We show qualitative examples in supplementary material and results of traditional guided matching on this dataset.

\subsection{Application to challenging 3D reconstruction}


We demonstrate that our approach can help 3D reconstruction in its most challenging cases by performing 3D reconstruction on the LTLL dataset~\cite{fernando2015location}. This dataset contains 25~sets of historical and recent pictures of the same scene. We tried to reconstruct the scenes both using historical images only and using recent and historical images. Many scenes are either too small or too complicated for 3D reconstruction with any method. We focus our analysis on the 5 scenes that D2-Net could reconstruct from historical images only, and the 8 scenes it could reconstruct from all photographs.


%

In Figure~\ref{fig:prop_registered}, we report the average proportion of registered images for the historical set using 4 different features with and without guidance from our network trained with epipolar supervision. For every feature the guided matching helps registering more images. In particular, guiding SIFT features with our methods registers the most images. Similar results for the reconstruction that uses all images are provided in supplementary material.
In Figure~\ref{fig:rec3D}, we show the example of reconstruction of the old set of ``Gare de Lyon'' scene for SuperPoint with and without guidance. Guiding SuperPoint features helps registering one more image and the obtained point cloud is way more complete.

\subsection{Limitations}

Our method has two main drawbacks.  First, 
it has a small but non negligible computational cost since the coarse matching adds an extra 70ms of computation time per image pair, to compare to the 30ms necessary to match 10000 SIFT keypoints. However it is dominated by the time of the 4D convolutions, which is currently based on loops of 3D convolution and could be made much faster by a direct CUDA implementation. 
Second, similar to traditional guided matching methods, it cannot be used to compute image visibility graphs for large scenes. Indeed, since they are trained on matching images, the coarse matches tend to be geometrically consistent even for input images representing different scenes. This limitation is not specific to our method : we provide quantitative evaluation and comparison with other methods in supplementary material.  
\section{Conclusion}

We have presented a new paradigm to perform local feature matching. Our key idea is to use a deep learning model to predict coarse matches between images and use them to guide classical feature matches. We discussed several possible supervisions for this coarse matching model, and demonstrated the benefits of a weak epipolar supervision. Our method boosts the performances that can be obtained with SIFT features to the level of recent learning-based features. We also showed it leads to state of the art results in 3D tasks such as visual localization and 3D reconstruction in challenging conditions. 

\paragraph{Acknowledgments:}
Fran\c{c}ois Darmon was supported by a CIFRE PhD grant from Thales LAS France and Mathieu Aubry by ANR project EnHerit ANR-17-CE23-0008.
We thank Bénédicte Bascle and Jean-Clément Devaux from Thales LAS for helpful discussions.

{\small
\bibliographystyle{ieee}
\bibliography{egbib}

\begin{thebibliography}{10}\itemsep=-1pt

\bibitem{bian2017gms}
J.~Bian, W.-Y. Lin, Y.~Matsushita, S.-K. Yeung, T.-D. Nguyen, and M.-M. Cheng.
\newblock {GMS}: Grid-based motion statistics for fast, ultra-robust feature
  correspondence.
\newblock In {\em Proceedings of the IEEE Conference on Computer Vision and
  Pattern Recognition}, pages 4181--4190, 2017.

\bibitem{bian2019bench}
J.-W. Bian, Y.-H. Wu, J.~Zhao, Y.~Liu, L.~Zhang, M.-M. Cheng, and I.~Reid.
\newblock An evaluation of feature matchers for fundamental matrix estimation.
\newblock In {\em British Machine Vision Conference (BMVC)}, 2019.

\bibitem{brachmann2019neural}
E.~Brachmann and C.~Rother.
\newblock Neural-guided {RANSAC}: Learning where to sample model hypotheses.
\newblock In {\em Proceedings of the IEEE International Conference on Computer
  Vision}, pages 4322--4331, 2019.

\bibitem{detone2016deep}
D.~DeTone, T.~Malisiewicz, and A.~Rabinovich.
\newblock Deep image homography estimation.
\newblock {\em arXiv preprint arXiv:1606.03798}, 2016.

\bibitem{detone2018superpoint}
D.~DeTone, T.~Malisiewicz, and A.~Rabinovich.
\newblock Superpoint: Self-supervised interest point detection and description.
\newblock In {\em Proceedings of the IEEE Conference on Computer Vision and
  Pattern Recognition Workshops}, pages 224--236, 2018.

\bibitem{dosovitskiy2015flownet}
A.~Dosovitskiy, P.~Fischer, E.~Ilg, P.~Hausser, C.~Hazirbas, V.~Golkov, P.~Van
  Der~Smagt, D.~Cremers, and T.~Brox.
\newblock Flownet: Learning optical flow with convolutional networks.
\newblock In {\em Proceedings of the IEEE international conference on computer
  vision}, pages 2758--2766, 2015.

\bibitem{dusmanu2019d2}
M.~Dusmanu, I.~Rocco, T.~Pajdla, M.~Pollefeys, J.~Sivic, A.~Torii, and
  T.~Sattler.
\newblock D2-net: A trainable {CNN} for joint description and detection of
  local features.
\newblock In {\em The IEEE Conference on Computer Vision and Pattern
  Recognition (CVPR)}, June 2019.

\bibitem{ebel2019beyond}
P.~Ebel, A.~Mishchuk, K.~M. Yi, P.~Fua, and E.~Trulls.
\newblock Beyond cartesian representations for local descriptors.
\newblock In {\em Proceedings of the IEEE International Conference on Computer
  Vision}, pages 253--262, 2019.

\bibitem{feng2011feature}
T.~Feng and J.~Yuan.
\newblock Feature point detection and matching of wide baseline image based on
  scale space theory and guided matching algorithm.
\newblock In {\em 2011 International Conference on Multimedia Technology},
  pages 538--542. IEEE, 2011.

\bibitem{fernando2015location}
B.~Fernando, T.~Tommasi, and T.~Tuytelaars.
\newblock Location recognition over large time lags.
\newblock {\em Computer Vision and Image Understanding}, 139:21--28, 2015.

\bibitem{fischler1981random}
M.~A. Fischler and R.~C. Bolles.
\newblock Random sample consensus: a paradigm for model fitting with
  applications to image analysis and automated cartography.
\newblock {\em Communications of the ACM}, 24(6):381--395, 1981.

\bibitem{geiger2012we}
A.~Geiger, P.~Lenz, and R.~Urtasun.
\newblock Are we ready for autonomous driving? the {KITTI} vision benchmark
  suite.
\newblock In {\em 2012 IEEE Conference on Computer Vision and Pattern
  Recognition}, pages 3354--3361. IEEE, 2012.

\bibitem{hartley2003multiple}
R.~Hartley and A.~Zisserman.
\newblock {\em Multiple View Geometry in Computer Vision}.
\newblock Cambridge university press, 2003.

\bibitem{he2016deep}
K.~He, X.~Zhang, S.~Ren, and J.~Sun.
\newblock Deep residual learning for image recognition.
\newblock In {\em Proceedings of the IEEE conference on computer vision and
  pattern recognition}, pages 770--778, 2016.

\bibitem{heinly2015reconstructing}
J.~Heinly, J.~L. Schonberger, E.~Dunn, and J.-M. Frahm.
\newblock Reconstructing the world *in six days* (as captured by the yahoo 100
  million image dataset).
\newblock In {\em Proceedings of the IEEE Conference on Computer Vision and
  Pattern Recognition}, pages 3287--3295, 2015.

\bibitem{ilg2017flownet}
E.~Ilg, N.~Mayer, T.~Saikia, M.~Keuper, A.~Dosovitskiy, and T.~Brox.
\newblock Flownet 2.0: Evolution of optical flow estimation with deep networks.
\newblock In {\em Proceedings of the IEEE conference on computer vision and
  pattern recognition}, pages 2462--2470, 2017.

\bibitem{knapitsch2017tanks}
A.~Knapitsch, J.~Park, Q.-Y. Zhou, and V.~Koltun.
\newblock Tanks and temples: {B}enchmarking large-scale scene reconstruction.
\newblock {\em ACM Transactions on Graphics (ToG)}, 36(4):1--13, 2017.

\bibitem{li2018megadepth}
Z.~Li and N.~Snavely.
\newblock Megadepth: Learning single-view depth prediction from internet
  photos.
\newblock In {\em Proceedings of the IEEE Conference on Computer Vision and
  Pattern Recognition}, pages 2041--2050, 2018.

\bibitem{lin2017code}
W.-Y. Lin, F.~Wang, M.-M. Cheng, S.-K. Yeung, P.~H. Torr, M.~N. Do, and J.~Lu.
\newblock {CODE}: Coherence based decision boundaries for feature
  correspondence.
\newblock {\em IEEE transactions on pattern analysis and machine intelligence},
  40(1):34--47, 2017.

\bibitem{lowe2004distinctive}
D.~G. Lowe.
\newblock Distinctive image features from scale-invariant keypoints.
\newblock {\em International journal of computer vision}, 60(2):91--110, 2004.

\bibitem{luo2019contextdesc}
Z.~Luo, T.~Shen, L.~Zhou, J.~Zhang, Y.~Yao, S.~Li, T.~Fang, and L.~Quan.
\newblock Contextdesc: Local descriptor augmentation with cross-modality
  context.
\newblock In {\em Proceedings of the IEEE Conference on Computer Vision and
  Pattern Recognition}, pages 2527--2536, 2019.

\bibitem{luo2018geodesc}
Z.~Luo, T.~Shen, L.~Zhou, S.~Zhu, R.~Zhang, Y.~Yao, T.~Fang, and L.~Quan.
\newblock Geodesc: Learning local descriptors by integrating geometry
  constraints.
\newblock In {\em Proceedings of the European Conference on Computer Vision
  (ECCV)}, pages 168--183, 2018.

\bibitem{ma2019locality}
J.~Ma, J.~Zhao, J.~Jiang, H.~Zhou, and X.~Guo.
\newblock Locality preserving matching.
\newblock {\em International Journal of Computer Vision}, 127(5):512--531,
  2019.

\bibitem{maier2016guided}
J.~Maier, M.~Humenberger, M.~Murschitz, O.~Zendel, and M.~Vincze.
\newblock Guided matching based on statistical optical flow for fast and robust
  correspondence analysis.
\newblock In {\em European Conference on Computer Vision}, pages 101--117.
  Springer, 2016.

\bibitem{melekhov2019dgc}
I.~Melekhov, A.~Tiulpin, T.~Sattler, M.~Pollefeys, E.~Rahtu, and J.~Kannala.
\newblock {DGC-Net}: Dense geometric correspondence network.
\newblock In {\em 2019 IEEE Winter Conference on Applications of Computer
  Vision (WACV)}, pages 1034--1042. IEEE, 2019.

\bibitem{mikolajczyk2005performance}
K.~Mikolajczyk and C.~Schmid.
\newblock A performance evaluation of local descriptors.
\newblock {\em IEEE transactions on pattern analysis and machine intelligence},
  27(10):1615--1630, 2005.

\bibitem{mikolajczyk2005comparison}
K.~Mikolajczyk, T.~Tuytelaars, C.~Schmid, A.~Zisserman, J.~Matas,
  F.~Schaffalitzky, T.~Kadir, and L.~Van~Gool.
\newblock A comparison of affine region detectors.
\newblock {\em International journal of computer vision}, 65(1-2):43--72, 2005.

\bibitem{mishchuk2017working}
A.~Mishchuk, D.~Mishkin, F.~Radenovic, and J.~Matas.
\newblock Working hard to know your neighbor's margins: {L}ocal descriptor
  learning loss.
\newblock In {\em Advances in Neural Information Processing Systems}, pages
  4826--4837, 2017.

\bibitem{mishkin2018repeatability}
D.~Mishkin, F.~Radenovic, and J.~Matas.
\newblock Repeatability is not enough: Learning affine regions via
  discriminability.
\newblock In {\em Proceedings of the European Conference on Computer Vision
  (ECCV)}, pages 284--300, 2018.

\bibitem{moo2018learning}
K.~Moo~Yi, E.~Trulls, Y.~Ono, V.~Lepetit, M.~Salzmann, and P.~Fua.
\newblock Learning to find good correspondences.
\newblock In {\em Proceedings of the IEEE Conference on Computer Vision and
  Pattern Recognition}, pages 2666--2674, 2018.

\bibitem{ono2018lf}
Y.~Ono, E.~Trulls, P.~Fua, and K.~M. Yi.
\newblock {LF-Net}: Learning local features from images.
\newblock In {\em Advances in Neural Information Processing Systems}, pages
  6234--6244, 2018.

\bibitem{plotz2018neural}
T.~Pl{\"o}tz and S.~Roth.
\newblock Neural nearest neighbors networks.
\newblock In {\em Advances in Neural Information Processing Systems}, pages
  1087--1098, 2018.

\bibitem{ranftl2018deep}
R.~Ranftl and V.~Koltun.
\newblock Deep fundamental matrix estimation.
\newblock In {\em Proceedings of the European Conference on Computer Vision
  (ECCV)}, pages 284--299, 2018.

\bibitem{ranjan2017optical}
A.~Ranjan and M.~J. Black.
\newblock Optical flow estimation using a spatial pyramid network.
\newblock In {\em Proceedings of the IEEE Conference on Computer Vision and
  Pattern Recognition}, pages 4161--4170, 2017.

\bibitem{revaud2019r2d2}
J.~Revaud, P.~Weinzaepfel, C.~De~Souza, and M.~Humenberger.
\newblock {R2D2}: Reliable and repeatable detector and descriptor.
\newblock In {\em Advances in Neural Information Processing Systems}, pages
  12405--12415, 2019.

\bibitem{rocco2017convolutional}
I.~Rocco, R.~Arandjelovic, and J.~Sivic.
\newblock Convolutional neural network architecture for geometric matching.
\newblock In {\em Proceedings of the IEEE Conference on Computer Vision and
  Pattern Recognition}, pages 6148--6157, 2017.

\bibitem{rocco2018end}
I.~Rocco, R.~Arandjelovi{\'c}, and J.~Sivic.
\newblock End-to-end weakly-supervised semantic alignment.
\newblock In {\em Proceedings of the IEEE Conference on Computer Vision and
  Pattern Recognition}, pages 6917--6925, 2018.

\bibitem{rocco2018neighbourhood}
I.~Rocco, M.~Cimpoi, R.~Arandjelovi{\'c}, A.~Torii, T.~Pajdla, and J.~Sivic.
\newblock Neighbourhood consensus networks.
\newblock In {\em Advances in Neural Information Processing Systems}, pages
  1651--1662, 2018.

\bibitem{sarlin2020superglue}
P.-E. Sarlin, D.~DeTone, T.~Malisiewicz, and A.~Rabinovich.
\newblock Superglue: Learning feature matching with graph neural networks.
\newblock In {\em Proceedings of the IEEE/CVF Conference on Computer Vision and
  Pattern Recognition}, pages 4938--4947, 2020.

\bibitem{sattler2018benchmarking}
T.~Sattler, W.~Maddern, C.~Toft, A.~Torii, L.~Hammarstrand, E.~Stenborg,
  D.~Safari, M.~Okutomi, M.~Pollefeys, J.~Sivic, et~al.
\newblock Benchmarking {6DOF} outdoor visual localization in changing
  conditions.
\newblock In {\em Proceedings of the IEEE Conference on Computer Vision and
  Pattern Recognition}, pages 8601--8610, 2018.

\bibitem{schonberger2016structure}
J.~L. Sch\"onberger and J.-M. Frahm.
\newblock Structure-from-motion revisited.
\newblock In {\em Proceedings of the IEEE Conference on Computer Vision and
  Pattern Recognition}, pages 4104--4113, 2016.

\bibitem{shah2015geometry}
R.~Shah, V.~Srivastava, and P.~Narayanan.
\newblock Geometry-aware feature matching for structure from motion
  applications.
\newblock In {\em 2015 IEEE Winter Conference on Applications of Computer
  Vision}, pages 278--285. IEEE, 2015.

\bibitem{sturm2012benchmark}
J.~Sturm, N.~Engelhard, F.~Endres, W.~Burgard, and D.~Cremers.
\newblock A benchmark for the evaluation of {RGB-D SLAM} systems.
\newblock In {\em 2012 IEEE/RSJ International Conference on Intelligent Robots
  and Systems}, pages 573--580. IEEE, 2012.

\bibitem{sun2018pwc}
D.~Sun, X.~Yang, M.-Y. Liu, and J.~Kautz.
\newblock {PWC-Net}: {CNNs} for optical flow using pyramid, warping, and cost
  volume.
\newblock In {\em Proceedings of the IEEE Conference on Computer Vision and
  Pattern Recognition}, pages 8934--8943, 2018.

\bibitem{sun2019attentive}
W.~Sun, W.~Jiang, E.~Trulls, A.~Tagliasacchi, and K.~M. Yi.
\newblock Attentive context normalization for robust permutation-equivariant
  learning, 2019.

\bibitem{taira2019inloc}
H.~{Taira}, M.~{Okutomi}, T.~{Sattler}, M.~{Cimpoi}, M.~{Pollefeys},
  J.~{Sivic}, T.~{Pajdla}, and A.~{Torii}.
\newblock Inloc: Indoor visual localization with dense matching and view
  synthesis.
\newblock {\em IEEE Transactions on Pattern Analysis and Machine Intelligence},
  pages 1--1, 2019.

\bibitem{thomee2016yfcc100m}
B.~Thomee, D.~A. Shamma, G.~Friedland, B.~Elizalde, K.~Ni, D.~Poland, D.~Borth,
  and L.-J. Li.
\newblock {YFCC100M}: The new data in multimedia research.
\newblock {\em Communications of the ACM}, 59(2):64--73, 2016.

\bibitem{imw}
E.~Trulls, Y.~Jin, K.~Yi, D.~Mushkin, J.~Matas, A.~Mishchuk, and P.~Fua.
\newblock Image matching benchmark, 2020.

\bibitem{widya2018structure}
A.~R. Widya, A.~Torii, and M.~Okutomi.
\newblock Structure from motion using dense {CNN} features with keypoint
  relocalization.
\newblock {\em IPSJ Transactions on Computer Vision and Applications}, 10(1):6,
  2018.

\bibitem{wilson2014robust}
K.~Wilson and N.~Snavely.
\newblock Robust global translations with {1DSFM}.
\newblock In {\em European Conference on Computer Vision}, pages 61--75.
  Springer, 2014.

\bibitem{xiao2013sun3d}
J.~Xiao, A.~Owens, and A.~Torralba.
\newblock Sun3d: A database of big spaces reconstructed using {SfM} and object
  labels.
\newblock In {\em Proceedings of the IEEE International Conference on Computer
  Vision}, pages 1625--1632, 2013.

\bibitem{yi2016lift}
K.~M. Yi, E.~Trulls, V.~Lepetit, and P.~Fua.
\newblock {LIFT}: Learned invariant feature transform.
\newblock In {\em European Conference on Computer Vision}, pages 467--483.
  Springer, 2016.

\bibitem{zhang2019learning}
J.~Zhang, D.~Sun, Z.~Luo, A.~Yao, L.~Zhou, T.~Shen, Y.~Chen, L.~Quan, and
  H.~Liao.
\newblock Learning two-view correspondences and geometry using order-aware
  network.
\newblock In {\em Proceedings of the IEEE International Conference on Computer
  Vision}, pages 5845--5854, 2019.

\end{thebibliography}
}

\end{document}